\def\year{2020}\relax
\documentclass[letterpaper]{article} 
\usepackage{aaai20}  
\usepackage{times}  
\usepackage{helvet} 
\usepackage{courier}  
\usepackage[hyphens]{url}  
\usepackage{graphicx} 
\usepackage{graphicx}  
\usepackage{todo}
\usepackage{amsmath}

\title{Deep RL Agent for a Real-Time Action Strategy Game}
\author{Michal Warchalski,\textsuperscript{\rm 1} Dimitrije Radojevic,
\textsuperscript{\rm 1} Milos Milosevic\textsuperscript{\rm 1}\\ 
\textsuperscript{\rm 1}Nordeus\\ 
11 Milutina Milankovica\\
11070 Belgrade, Serbia\\
\{michalw, dimitrijer, milosmi\}@nordeus.com 
}

\begin{document}

\maketitle

\begin{abstract} 
	We introduce a reinforcement learning environment based on
	Heroic - Magic Duel, a 1 v 1 action strategy game. This domain is non-trivial for
	several reasons: it is a real-time game, the state space is large, the
	information given to the player before and at each step of a match is
	imperfect, and distribution of actions is dynamic. 

	Our main contribution is a deep reinforcement learning agent
	playing the game at a competitive level that we trained using PPO and
	self-play with multiple competing
	agents, employing only a simple reward of $\pm 1$ depending on the outcome
	of a single match. Our best self-play agent, obtains around $65\%$
	win rate against the existing AI and over $50\%$ win rate against a top human
	player.
\end{abstract}

\section{Introduction}

Deep reinforcement learning for games has become an exciting area of research
in the last years, proving successful, among others, in Atari games
\cite{mnih2013playing,mnih2015humanlevelct}, in the games of Go and chess
\cite{silver2016masteringtg,silver2017mastering,silver2017masteringtg,silver2018agr}
leading to major progress in the complex real-time game environments of Dota
2 \cite{openai2019dota} and StarCraft 2 \cite{vinyals2017starcraft,vinyals2019grandmasterli}.

Inspired by the recent advances, we tackle a new reinforcement learning
environment based on the video game Heroic - Magic
Duel\footnote{www.heroicgame.com}. The game combines elements of a card game
and a real-time strategy game. Heroic has been downloaded over a million times
since it launched in June 2019, and more than half a million matches are being
played per day at the time of writing.

The players create custom decks consisting of cards that they will play against
each other in a live match which usually lasts between 2 and 3 minutes. It is
played 1 v 1, in real-time, on a symmetrical battlefield that separates your
castle from the opponent's. Players take actions by playing cards in
order to deploy units on the battlefield, which fight the opponent's units with
the goal of destroying their castle.
In addition to deploying units, players have an additional action available to
them - casting spells. These usually affect units in some way and are purely
limited by the fact that once you use them - they cannot be cast again for a
period of time (they are on ``cooldown''). In addition to this, deploying
units uses a specific resource (mana) which refreshes over time - meaning that
if a certain unit is deployed, another one cannot be deployed for a certain
period afterwards.

There are several reasons why this domain is non-trivial. First of all, it is a
real-time game with a considerable state space, since there is a number
of different units in the game and coordinates on every lane are expressed in
floating point precision. Secondly, the information given is imperfect both
before and at the time of the game. More concretely, the decks that both
players use in a match are a priori unknown and at any time of the match neither
player knows which cards are currently available to the opponent.
Moreover, because of the aforementioned cooldown and mana, taking any action has
an intrinsic cost of limiting further action use for a time. 

\begin{figure}[t] \centering
    \includegraphics[width=\columnwidth]{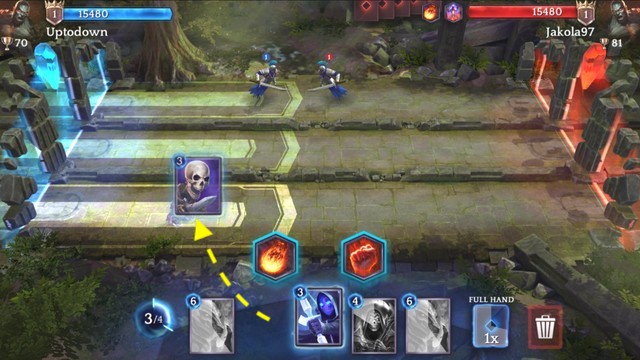}
\caption{Game screen. The yellow dashed arrow represents the card casting mechanics.}
\label{figure1}
\end{figure}

As the main contribution of this paper, we trained a deep reinforcement learning
agent playing the game at a competitive level via self-play and proximal policy
optimization algorithm \cite{schulman2017proximal}, using only a simple reward
of $\pm 1$ depending on the outcome of a single match. The agent is robust to
both different strategies and decks used. Following \cite{bansal2017emergent} in
the learning process the agent is competing against an ensemble of policies
trained in parallel and, moreover, before every match the decks of both the
agent and the opponent are randomly sampled from a pool of decks. With such
distribution of decks our best self-play agent has achieved around $65\%$ win
rate against the existing AI and $50\%$ win rate against a top human player.

In the following we further elaborate on the parts that we found crucial for our
work.

The game offers plenty of visual
information, which makes it challenging to learn a policy from pure pixels.
Similarly to \cite{vinyals2017starcraft}, we represent the spatial state of the
game by constructing low resolution grids of features, which
contain positions of different units on the map and use it together with the
non-spatial information as the observation. 

As for the value network we adopt
the Atari-net architecture appearing in \cite{vinyals2017starcraft} in the
context of StarCraft 2. Regarding the policy network we initially started with
the same baseline, which did not give us good results. We upgraded the policy by
adding more flexibility to spatial actions. The change is twofold.
First, we condition the spatial action on the selected action in the
network head instead of sampling it independently of
the selected action. Second, we output a single spatial action instead
of sampling its coordinates separately. We found that these modifications
improved the results by a substantial margin. In their implementation we make
use of the fact that the action space and the discretized representation of the
map is not very large. 

Through experiments we found that the version of the game that involves
spells is considerably more complex. Our hypothesis is that this is because they
are usually required to be cast with more precision and it is often more desired 
to save a spell for later than to cast it immediately.

We test several training curricula including playing against existing AI and
via self-play, where the agent is trained against an ensemble of policies that
are being trained in parallel \cite{bansal2017emergent}. To measure the
performance we test the agent against two AI systems: rule-based and
tree-search based. We also measure performance of the agent against a top $95$th
percentile human
player.

It is crucial to skip meaningless no-op actions in the experiments. Most of the
time during a match no action is available, which makes no-op action useless
from the strategical point of view and impacts the learning. We resolve this
issue by returning the state only when there is a non-trivial action available. 

\section{Related Work}
In this section we elaborate on the work related to the paper, namely, deep
reinforcement learning agents, proximal policy optimization, self-play,
representation of spatial information in games as feature layers and passive
no-op actions.

There has been a lot of progress in the area of deep reinforcement learning in
games which started several years ago with agents obtaining superhuman level
of performance in arcade games \cite{mnih2015humanlevelct,mnih2016asynchronous}. Such
approach, combined with tree-search techniques and self-play, also proved
successful in complex board games \cite{silver2016masteringtg,silver2017mastering,silver2017masteringtg,silver2018agr}.
Most recently deep reinforcement learning agents playing real-time strategy
games beat top professional human players 
\cite{openai2019dota,vinyals2019grandmasterli}.

Proximal policy optimization (PPO) introduced in \cite{schulman2017proximal} is a
popular on-policy algorithm in the reinforcement learning community due to its
performance and simplicity. It was successfully used recently for training a
neural network that beat top Dota 2 players \cite{openai2019dota}. PPO is often
combined with generalized advantage estimation (GAE) from \cite{schulman2015highdimensional}.

Deep reinforcement learning agents for competitive games are usually trained
via self-play \cite{silver2018agr,openai2019dota} or combining self-play with
imitation learning \cite{vinyals2019grandmasterli}. In order to provide a
variety of strategies, it has become popular to train multiple competing agents
in parallel \cite{bansal2017emergent,vinyals2019grandmasterli}.

In order to simplify the input of policy and value networks while still
preserving the crucial spatial information, \cite{vinyals2017starcraft} use low
resolution grids of features corresponding, among other things, to positions of
various types of units on the game map and their properties. Such grids are
then fed into a convolutional layer of a neural network in the form of a tensor
with multiple channels together with the non-spatial information.
Subsequently, the processed feature layers and the non-spatial component are
combined (for example via flattening and concatenation) and fed into the head
of the policy network, which outputs a probability distribution over actions.
It is also worth noting that feature layers representing the positions of chess
pieces at several past timesteps were also used in 
\cite{silver2017mastering,silver2017masteringtg}.

The authors of \cite{oh2019creating} encountered a related
issue concerning no-op actions, when no other action can be executed.
Although they tackle a quite different problem, namely a real-time fighting
game, similarly as in our case they resolve it by skipping passive no-op
actions.

\section{The Heroic Environment} 

In this section we present details of the environment based on Heroic -
Magic Duel. In particular, we describe the game itself
and introduce the observation and the action space of the reinforcement learning
environment used throughout the paper.
Finally, we give some technical details on the implementation of the system.

\subsection{Game Details}

\subsubsection{Overview of the Game} 

The map where the match happens fits onto one screen of a mobile phone, see
Figure~\ref{figure1}. Map is symmetrical, consisting of a battlefield which
separates two castles, blue belonging to the left player, and red belonging to
the right player. It is itself split into 3 separate lanes. Lanes are separate
parallel linear battlefields which all start at one castle and end at the
other. All lanes have access to the enemy castle at the end of
it. The goal is to destroy the enemy castle.

\begin{figure}[h] \centering
    \includegraphics[width=\columnwidth]{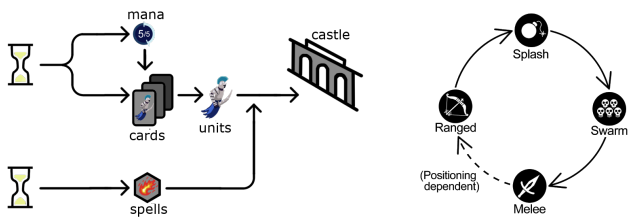}
\caption{Schematic diagram of the game mechanics (left) and diagram of attack dependencies (right).}
\label{mana_and_diagram}
\end{figure}

\subsubsection{Cards and Units} 

Before the match, the player selects a deck of 12 cards to be used during the
match out of the total of 56 possible cards that exist in the game. At the time
of the match, players play their cards at a desired location on the map,
effectively spawning the units there.

\subsubsection{Card Action Casting} During the match, the players will spend
mana to play cards from their hand in order to cast units onto the battlefield.
They start the match with 4 cards in their hand, and periodically draw one more
card to refill it from the deck. Player's current hand is represented as cards
at the bottom of the screen, see Figure~\ref{figure1}. The deck holds an
infinite supply of cards in the ratio that is determined by the 12 cards they
previously selected. In order to be able to play a certain card, they will need
to pay the cost in mana. The mana is a resource which slowly refills up until it
reaches the mana cap - which is the limit of mana supply the current player can
hold.
Each player starts the match with no mana, and a mana cap of 2. Mana cap
increases
at the same pace as drawing of cards does - meaning that more expensive cards
can be played only at later stages of the match. Due to the fact that mana can
be capped (and thus not refill until it is spent by playing of a card), and that
hand can be capped by a maximum of 5 cards - (and thus not refill with a new
card, until one is spent), there is a pressure to keep playing actions in order
to use the steady flow of resources optimally, see Figure~\ref{mana_and_diagram}.

To play a card, players drag it from their hand to the position where they
want to cast it on the map. Once the card is cast, the unit corresponding to
the card appears and the card is removed from the hand. The unit automatically
starts moving towards the opponent's castle (e.g.\ if the left player casts the
card, the unit will start moving towards the right end of the lane). A card can
be cast only between one's own castle and the closest opponent unit on the
given lane, but no further than the furthest friendly unit on any of the
3 lanes, and no further than half-length of the lane. This means at the start of
the match players can only cast units at their castle.

For example, in Figure~\ref{figure1}, the left player is playing his card by
dragging it on the bottom lane. Additionally, he has four other cards in his
hand, one of them is fully available, while the other three require more mana.

\subsubsection{Unit Intrinsic Behavior} 
A unit on the battlefield automatically moves towards the opponent castle. If
it manages to get to the end of the lane it starts attacking the opponent
castle,
lowering its health points. If any player's castle hit points reach 0, that
player loses the match, and their opponent wins the match. Units can only move
within, and interact with enemies that are in the same lane as that unit. If the
moving unit encounters an opposing unit on the way to the enemy castle it
will start attacking this opposing unit, attempting to destroy it by depleting
it of health points in order to proceed towards the goal - this is how units
``fight'' each other. Speed and fighting abilities are determined by the game (and
specified on
the card that was used to cast that unit) - the player cannot influence any of
them after the card is cast. Fighting abilities of the unit are determined by a
number properties, such as its range, type of attack, and any special abilities
it might have. Certain unit types are good against certain types of units, but
lose to others. This gives the choice of which units to cast the dimension of a
rock-paper-scissors problem.

For example, Figure~\ref{mana_and_diagram} shows that units that have a splash attack -
attack which hits multiple enemies at once - are good against units which
consist of
many weak enemies, which are the ``swarm'' archetype. These are strong against
melee type - the most basic units which attack one enemy at a time in close
range. Melee type is normally good against ranged units - but this interaction
can depend on how each unit is placed. Ranged units are again good against
splash units as splash units are generally slow moving, and do not do great
against single targets. Additionally, since ranged and swarm units have low hit
points (they are easily defeated when attacked), they are generally weak against
aggressive spells.

\subsubsection{Spells} In addition to preparing their decks of cards, players
choose 2 spells out of a total of 25 to be available to them to use during the
match. Spells are special actions that require no mana to be played, but once
played cannot be cast for a set period of time - the spell is on cooldown.
Spells have a variety of effects, usually affecting units in some way. For
example, they can lower the opponent units' health points in an area where they
are cast or make one's own units stronger. Spell actions are generally more
powerful but less frequently available than casting unit actions, and correct
use can swing the match in one's favor. The spells that are available to be
cast from the interface are represented as hexagons above the player's hand of
cards on the game screen, see Figure~\ref{figure1}.

\subsubsection{Spell Action Casting} The player can select any of the available
spells and place it at any location on the map. Once the spell is cast, the
special action happens, which lasts up to couple of seconds. Spells take effect
only in a certain range. There are no restrictions on when a spell can be cast,
but once it is, it is on cooldown - meaning it cannot be cast again for a
period of 25 - 60 seconds, the length of the cooldown is the property of the
spell itself and is enforced by the game, see Figure~\ref{mana_and_diagram}.

\subsection{Reinforcement Learning Environment}

In this subsection we present specifications of the reinforcement learning
environment that we used in the experiments. We start by introducing some
necessary setup.

Let $U=56$ be the total number of units in the game, let
$L=3$ be the number of lanes. We discretize each lane by splitting it into
$D=10$ bins of equal length. Let $S=25$ be the
number of available spells. Let $A = U + S + 1$, be the total number of actions
in the game,
i.e.\ the total number of units (since it is equal to the number of cards), the
total number of spells plus no-op action.
We index the unit type, lanes, discretization bins on every lane and the
actions with the sets\footnote{For any natural number $n$, we set $[n] := \{0,1,
\ldots, n-1\}$.} $[U]$, $[L]$, $[D]$ and $[A]$,
respectively.

\subsubsection{Observations} We represent an observation as pair $(o_s,
o_{ns})$, where $o_s$ is the spatial and $o_{ns}$ the non-spatial component,
respectively.

The spatial component $o_s$ is a tensor of shape $D \times L \times 2U$ where
for $u \in [U]$ the value of $o_s(d, l, u)$ is the sum of health (in percent) of
own units of type $u$ on lane $l$ whose position on the lane falls into the
$d$-th bin of the discretization and for $u \in [2U] \setminus [U]$ it is the
sum of health of opponent's units of type $u - U$ at the same $d$, $l$.

The non-spatial component $o_{ns}$ is a vector of length $A + 3$, where for $a
\in [A]$, $o_{ns}(a)$ is the number of seconds till the action $a$ is
available. If this time is not known, there are two possibilities. The first one
is that the action is available in the match, but the time-till-available is
unknown. For such actions we set a default large positive value. The other
possibility is that the action is not available in the current match, in this
case we set a default negative value. The last three coordinates of $o_{ns}$
correspond to own castle health, opponent's castle health and elapsed match
time, respectively.

\subsubsection{Actions} We represent an action as triple $(a, x, y)$, where $a
\in [A]$ is the selected action type and $y \in [L]$, $x \in [D]$ are the lane
and the position on the lane where the action should be executed. We refer to 
$a$ as the non-spatial action and $(x, y)$ as the spatial action. Note that in
the case of no-op action $x, y$ coordinates are
irrelevant. Nevertheless, we do not give this action any special treatment.

\subsubsection{Reward} Throughout the experiments we give the agent reward of
$+1$ if it wins and $-1$ if it gets defeated. We also use discounted returns
with parameter $0 \leq \gamma \leq 1$.

\subsubsection{Technical Details} In this part we provide some details of the
implementation of the match simulation. Heroic - Magic Duel game has clear
separation of game view, which is used for displaying the match, and
simulation, which is used to simulate matches. We extract simulation part to a
service that is able to run multiple matches concurrently. Given that match
mechanics are deterministic, i.e.\ a discrete time step is used, it is possible
to simulate the entire match in only a fraction of real match time.  During
training, agent sends its action to the service, then service applies given
action and simulates the match up to the point where agent can act again, and
finally returns observation back to the agent.

\section{Reinforcement Learning Agent}

In this section we introduce the policy and the value networks that we use as
well as their updates during training. 

Let $\mathcal{A} := [A]$ be the set of all possible non-spatial actions in the
game,
$\mathcal{A}_{\mathrm{card}} \subset \mathcal{A}$ be the set of card actions 
\textit{together with} no-op action and let
$\mathcal{A}_{\mathrm{spell}} \subset \mathcal{A}$ be the spell actions
\textit{together with} no-op action. Moreover, let $\mathcal{O}$ be the set
of all observations.

\subsection{Policy Network} 

\begin{figure}[h] \centering
    \includegraphics[width=\columnwidth]{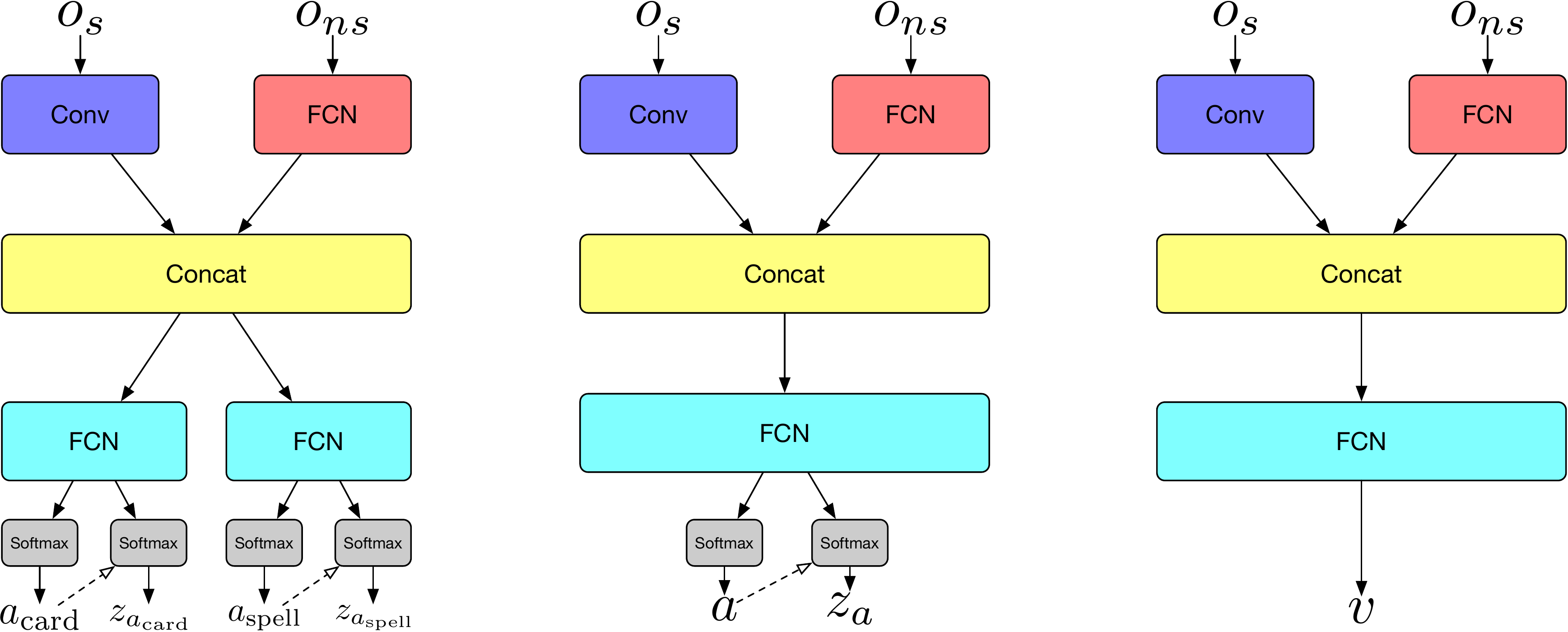}
    \caption{Two-headed policy network (left), single-headed policy network 
    (center) and value network (right). Both policies take an observation 
    $(o_{s}, o_{ns})$ consisting of a spatial and a non-spatial observation. $o_s$ is processed with one convolutional layer, while $o_{ns}$ is processed with a fully connected layer.
	The output of the convolutional layer is subsequently flattened and both vectors are
	concatenated. The result is then passed to a network head corresponding to a set $\mathcal{A}_* \in \{\mathcal{A}_{\mathrm{card}}, \mathcal{A}_{\mathrm{spell}}, \mathcal{A}\}$. It starts with a fully connected network. Next, a softmax layer that outputs a probability distribution over actions in $\mathcal{A}_*$ and a sequence of probability distributions of spatial actions, one for each $a \in \mathcal{A}_*$. Similarly to \cite{vinyals2017starcraft} we mask out invalid actions and renormalize the distribution. Next, an action $a \in
	\mathcal{A}_*$ is sampled. Finally, the spatial distribution
	corresponding to $a$ is used for sampling a spatial action $z_{a}$.}
    \label{policy_and_value}
\end{figure}

We model the policy as $\pi_\theta(\cdot|\cdot)
\colon \mathcal{A} \times \mathcal{O} \to [0, 1]$, where $\theta$ are the
parameters of a neural network. For every observation $o$ and action $a$, the
probability of taking $a$ given $o$ is given by $\pi_\theta(a|o)$. Below we
describe the architecture of the policy network.

Similarly to the Atari-net in \cite{vinyals2017starcraft} we process the
spatial observation $o_s$ with a convolutional network and the non-spatial
observation $o_{ns}$ with a fully connected network, whereupon we flatten the
output of the spatial branch and
concatenate both vectors. 

Concerning further stages of the architecture, we employ a two-headed 
and a single-headed policy network. In the following we describe a single policy head, which outputs
$a \in \mathcal{A}_*$ and a spatial action, where $\mathcal{A}_*$ is one of $\mathcal{A}_{\mathrm{card}}$,
$\mathcal{A}_{\mathrm{spell}}$, $\mathcal{A}$, depending on the architecture and the head considered.

A head outputs an action $a \in \mathcal{A}_{*}$, and an accompanying spatial
action. A difference between the Atari-net and our approach is that the output
spatial action $z_{a} := (x,y)_{a}$ is conditioned on the selected action $a$. 
We do it by outputting a distribution of spatial actions 
for each $a \in \mathcal{A}_{*}$, including invalid actions.
Subsequently, we mask out the invalid actions, sample a valid $a$ and use the spatial action distribution corresponding to $a$ in order to sample a sample spatial action $z_{a}$. Note that we can do that since the total number of cards and spells in the game is not very large, 
otherwise it could be costly to use multiple distributions with the same approach. Another difference is
that we output a single spatial action corresponding to a pair $(x,y)$, i.e.\ we do not sample $x$, $y$
independently as it is done in Atari-net. We found such approach natural, since products of probability
distributions form a smaller class of 2D probability distributions and, moreover, because $y$ is
intrinsically a discrete value. Here we make use of the fact that the ranges of both $x$, $y$ are not large,
which guarantees almost no impact on the size or the stability of the network.

\subsection{Value Network}

The main building blocks of the value network are the same as the ones of
Atari-net in \cite{vinyals2017starcraft} with the hyperparameters adjusted for our
problem. Similarly as in the policy network we feed the spatial observation
$o_s$ into a convolutional network and the non-spatial observation $o_{ns}$
into a fully connected network, subsequently we flatten the output of the convolutional
net and concatenate both outputs. Then, we pass the resulting vector to a fully
connected layer which outputs a single real value $v$, see Figure~\ref{policy_and_value}.

\subsection{Policy Network Update}

We train agents using proximal policy optimization (PPO) introduced in \cite{schulman2017proximal},
which is the state-of-the-art deep reinforcement learning algorithm in many
domains, including real-time games with large state and action spaces \cite{openai2019dota}. In this subsection we briefly describe it. 

Given policy $\pi_\theta(\cdot|\cdot)$, PPO update $\theta$ via
\begin{equation}
    \theta_{k+1} = \mathrm{arg}\,\mathrm{max}_{\theta} \,\,\, \mathbb{E}_{s,a \sim \pi_{\theta_k}}[L(s, a,\theta_k, \theta)],
    \label{ppo-update} 
\end{equation}
with
\begin{equation*}
L(s, a,\theta_k, \theta) := g\Big(\frac{\pi_\theta(a|s)}{\pi_{\theta_k}
(a|s)}, A^{\pi_{\theta_k}}(s, a), \epsilon\Big),
\end{equation*}
where $g(x,y,z) = \min\big(xy, \mathrm{clip}(x, 1-z, 1+z)y\big)$ and $A^
{\pi}$ is
the advantage function. The right hand side of Equation~\ref{ppo-update} is
estimated using the Monte-Carlo method. In the experiments we handle $A^{\pi}$
using generalized advantage estimate estimation (GAE) if the KL-divergence
between $\pi_{\theta_{k+1}}$ and $\pi_{\theta_k}$ becomes too large.

\subsection{Value Network Update}

PPO is coupled with learning the value network. We parametrize the value
network $V_\phi$ with $\phi$ and update it by minimizing the expected squared
error between $V_\phi(s_t)$ and the discounted
return $R_t = \sum_{t'\geq t} \gamma^{t' - t} R(s_{t'}, a_{t'})$, where $t$
denotes the timestep. More concretely, the update is as follows $$ \phi_k = 
\mathrm{arg} \, \mathrm{min}_{\phi} \,\,\, \mathbb{E}_{s_t, R_t \sim \pi_
{\theta_k}} [(V_\phi(s_t) - R_t)^2],$$ where the right hand side is estimated
via the Monte-Carlo method.

\section{Training Curriculum}

In this section we describe the curriculum that we use for training agents.
First, we specify the AI that we use as benchmarks to test the agents against.
Then, we elaborate on how we sample decks during learning.
Finally, we introduce two training procedures. The first one is based on playing
against the benchmark bots, while the second one is based on competing against
an ensemble of policies \cite{bansal2017emergent}.

\subsection{Benchmarking against Existing AI}
We use two different AI to benchmark performance of our agents. 
The first one is a rule-based AI, which utilizes multiple built-in
handcrafted conditions in order to decide which actions to execute.
The second one is a tree-search AI, which uses a tree
search algorithm combined with a handcrafted value function.

\subsection{Deck Sampling}
A single deck in the match consists of $12$ cards and may contain duplicates,
while in total there are $56$ cards in the game. In order to
make the agent robust under various decks we perform deck
sampling. Not all choices of decks are reasonable and in order to simplify the
selection process we prepared a pool of over $200$ decks chosen in collaboration
with a top player. In the training we uniformly sample one deck for the agent
and
one for the opponent.

Note that it would be an interesting problem to learn deck selection as a part
of agent's task, but we are not aiming to do it in this work.

\subsection{Training via Competing against AI} 
Given the existing AI that lets us perform fast rollouts, the simplest
curriculum is to simply pit an agent against a bot during training.
This has the advantage of quicker progress, which allowed for faster iterations.
The main disadvantage is the agent fitting its policy against a particular type
of opponent, which can result in low versatility.

\subsection{Training via Self-Play with Ensemble of Policies}

The second approach that we take is self-play, which recently became popular in
combination with deep reinforcement learning
\cite{silver2016masteringtg,silver2017mastering,silver2017masteringtg,silver2018agr,vinyals2019grandmasterli,openai2019dota}.
However, agents trained via simple self-play with one agent competing against older versions of itself tend to
find poor strategies and depending on the complexity of the environment there
are several ways to mitigate the issue \cite{bansal2017emergent,vinyals2019grandmasterli}.
In this paper we follow the strategy proposed in \cite{bansal2017emergent}
and train an ensemble of $n$ policies competing against each other. Every
several iterations we first sample two policies (could be the same policy) and
for each policy we select weights of the policy uniformly from the interval $
(\delta v, v)$, where $v$ is the last iteration number for the latest available
parameters and $0 \leq \delta < 1$ is a parameter of the algorithm, see 
\cite{bansal2017emergent} for details.

\section{Experiments}

We train deep reinforcement learning agents using the architectures and the
training plans introduced in the previous sections. 
Our goal is to demonstrate that these methods can yield agents playing the game
at a competitive level. 
Concerning the training plans we demonstrate with
experiments that the self-play curriculum yields agents with versatile strategies
against different types of opponents. We test the self-play agents against the existing AI and a top human player. We also study the game with spells enabled.
We point out that this version of the game is considerably more complex than the restricted counterpart and we examine different ways of handling the associated actions via the introduced policies. Before we present the results, we make a remark on handling no-op actions.

\subsection{No-Op Actions}

In the early stage of experimentation we discovered that naively triggering the
network too often (every 0.2, 0.5 or 1 second), even when no move is available,
can result in poor strategies. This is because usually for the vast majority of
match time no action is available to be played. Similarly as in
\cite{oh2019creating} we mitigate this issue by letting the agent perform no-op
actions only when there is a non-trivial action available. The duration of
no-op is a hyperparameter of our learning procedure and in the experiments we
employ $4$ seconds for card no-op and $4$ seconds for spell no-op. 

\subsection{Experiment Details}
\subsubsection{Scaling}
We train the agents on 2 Nvidia Titan V GPUs. We scale training using
distributed PPO implementation with $12$ parallel processes. Each worker
collects multiple trajectories, and the data is then used to perform a
synchronized update step.

\subsubsection{Policy and Value Network}
We use the policy and the value network architectures introduced above. The
first two layers of all networks are the same: we use $32$ filters of size $(1,
3)$ and stride $(1, 1)$ in the convolutional layer and $32$ units in the fully
connected layer.
We use
$512$ units in the fully connected layers in the policy head as well as in the
last layer of the value network.
\subsubsection{Algorithm Parameters}
We use PPO algorithm with clip ratio $0.2$, policy network learning
rate $0.0003$, value network learning rate $0.001$. We estimate the value
function with the advantage estimator $\mathrm{GAE}(\gamma, \lambda)$ with
discount factor
$\gamma=0.99$
and parameter $\lambda=0.97$. In each iteration we
collect $60000$ samples, which is equivalent of approximately $3000$ matches,
and
perform several epochs of update with batches of size $2500$. We also employ
early stopping if the approximate mean KL-divergence between the current and the
previous policy becomes greater than $0.01$.

\subsubsection{Episode and Match Length} We noticed that collecting more samples
per iteration helped stabilize the training and improved exploration. We found
value of $60000$ to be a good trade-off between stability and training speed.
Matches without spells enabled took on average around $14$ steps and with spells
around $25$ steps.

\subsection{Effect of Card No-Op}
\begin{figure}[h] \centering
    \includegraphics[width=0.8\columnwidth]{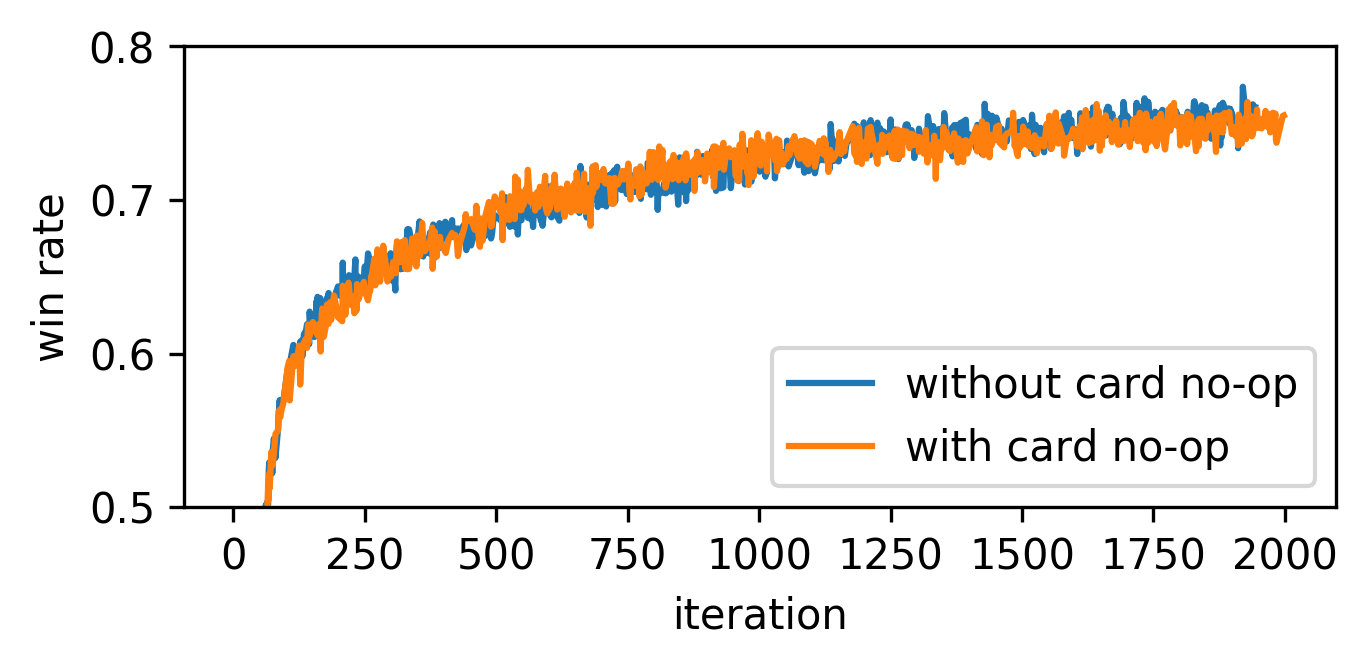}
\caption{Policy without vs with card no-op.} 
\label{effect-no-op-wo-spells}
\end{figure}

We compare two agents in the version of the game without spells - one with no-op
disabled and the other with no-op enabled. We train them against the rule-based
AI and use the policy in Figure~\ref{policy_and_value} (since spells are disabled, 
both introduced policies are equivalent).
In Figure~\ref{effect-no-op-wo-spells} we demonstrate the win rates of both
agents
over the course of training. They perform almost the same - our hypothesis is
that this is because the number of cards in player's hand is bounded (by $5$) and
new cards cannot be drawn as long as there is no free slot  -
hence, no-op actions do not give the agent any advantage and, as we additionally
observed during the experiment, the agent learns to avoid them over time. 

\subsection{Effect of Conditioning Spatial Actions} 
\begin{figure}[h] \centering
    \includegraphics[width=0.8\columnwidth]{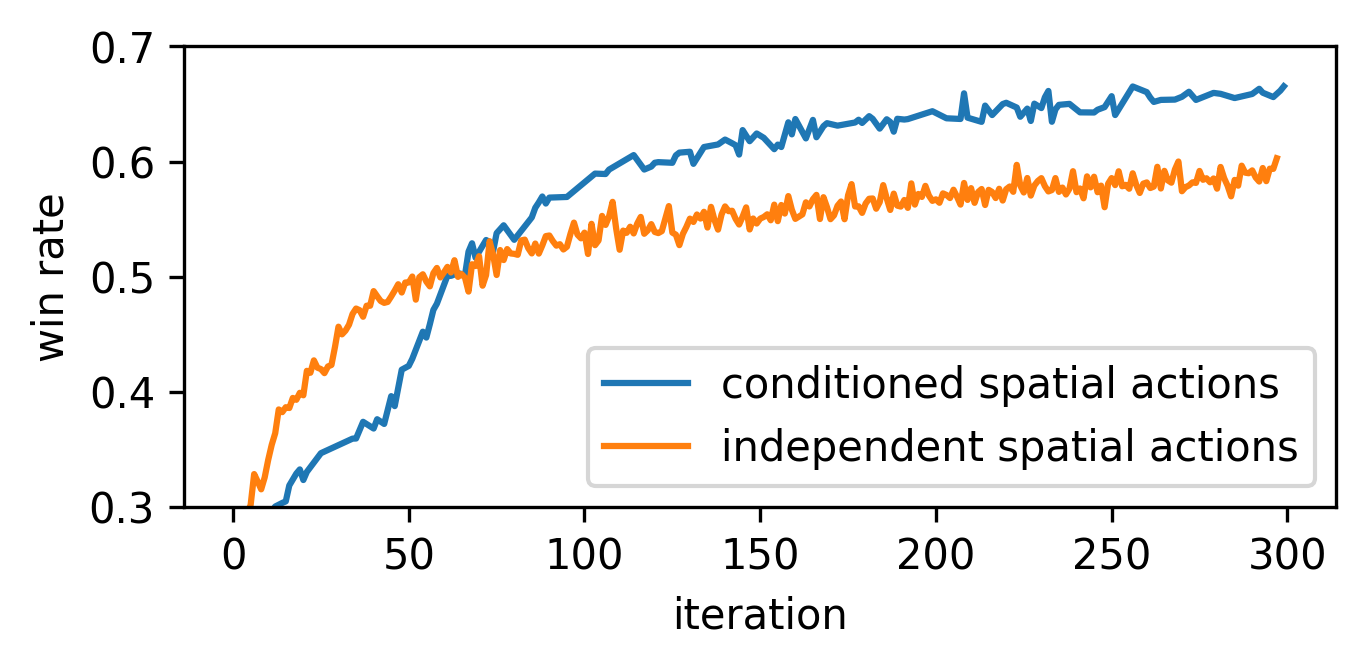}
\caption{Policy conditioned spatial actions vs policy
with spatial actions independent of the card action.} 
\label{effect-cond-spatial-actions}
\end{figure}

We compare our policy
network (Figure~\ref{policy_and_value}) with an Atari-net-like policy network.
The difference is that the latter samples spatial actions independently of the
selected card action and, moreover, samples $x$ independently of $y$. We run
experiments without spells and with no-op action disabled against the existing
rule-based AI.\ In Figure~\ref{effect-cond-spatial-actions} we demonstrate the
win rate of both agents in several hundred initial epochs - even in such
short training time the agent using conditioning of spatial actions performs better than
the other by a margin of about $10\%$. Note, however, that in the first
phase of training the performance of both agents is inverted - we think that
this is because the simpler agent discovers basic rules of casting cards more
quickly, while the other agent needs more time to do so.

\subsection{Effect of Self-Play}
We present the results of training agents via self-play with ensemble of
policies similarly as in \cite{bansal2017emergent}. We run the experiment with
$n=3$ policies and sampling interval $(\delta v, v)$ with $\delta=0.5$,
where $v$ is the number of the last iteration for a given agent. We periodically
tested the agents against the existing AI in order to track agents'
progress and versatility. We run the experiment without spells. Our best
self-play agent obtained $64.5\%$ against the rule-based AI and $37.3\%$ against
the tree-search AI. 

We also tested the agent against a human player, who is a developer that worked
on the game and is within top $95$th percentile of Heroic players. In total $20$
matches were played with random decks from the pool for both players. In such
test our self-play agent achieved $50\%$ win rate.

We additionally tested the self-play agent, the rule-based AI and the tree-search AI 
against the same human player in a series of $10$ matches against each of the three. 
This time the self-play agent obtained $60\%$ win rate, while the rule-based AI 
achieved $20\%$ win rate and the tree-search AI reached $50\%$ win rate.

\subsection{Effect of Enabling Spells}
\begin{figure}[h] \centering
    \includegraphics[width=0.8\columnwidth]{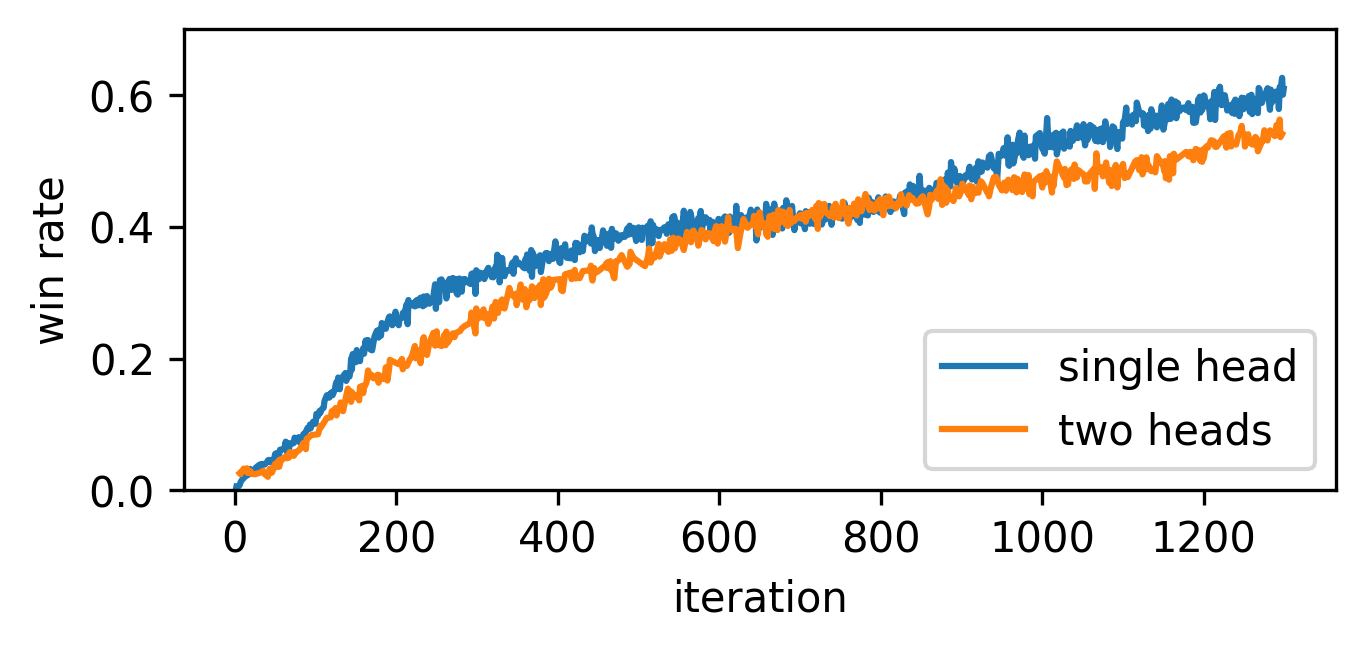}
\caption{Single-headed policy vs two-headed policy.} 
\label{effect_single_head_two_heads}
\end{figure}

In this subsection we study the effect of enabling spells in the game.
First we compare the single-headed and the two-headed policy network (Figure~\ref{policy_and_value}), which are two different ways of handling spells. 
We use the two-headed policy network with card
no-op disabled (see the card no-op experiment) and spell no-op enabled. The
single-headed policy uses a single
unified no-op action. In Figure~\ref{effect_single_head_two_heads} we
demonstrate the win rates of both agents. To our surprise, we observed that even
after a large number of iterations the win rate of the single-headed agent
remained better by a margin of a few percent. Note that there is a change of slope 
at a certain point (around the $800$-th iteration) of the training curve of the single-headed agent - we discovered that from this point on the agent started to use no-ops more often. 
In order to evaluate the importance of spell no-op actions, we ran additional experiments with both card no-op and spell no-op actions disabled or enabled. The results are indecisive - on one hand the performance of the agents was very similar to the performance of the agents in the previous experiment. On the other hand, 
as opposed to the experiments without spells (Figure~\ref{effect-no-op-wo-spells}), the frequency
with which the agents were using no-op actions did not drop during training.

\begin{figure}[h] \centering
    \includegraphics[width=0.8\columnwidth]
    {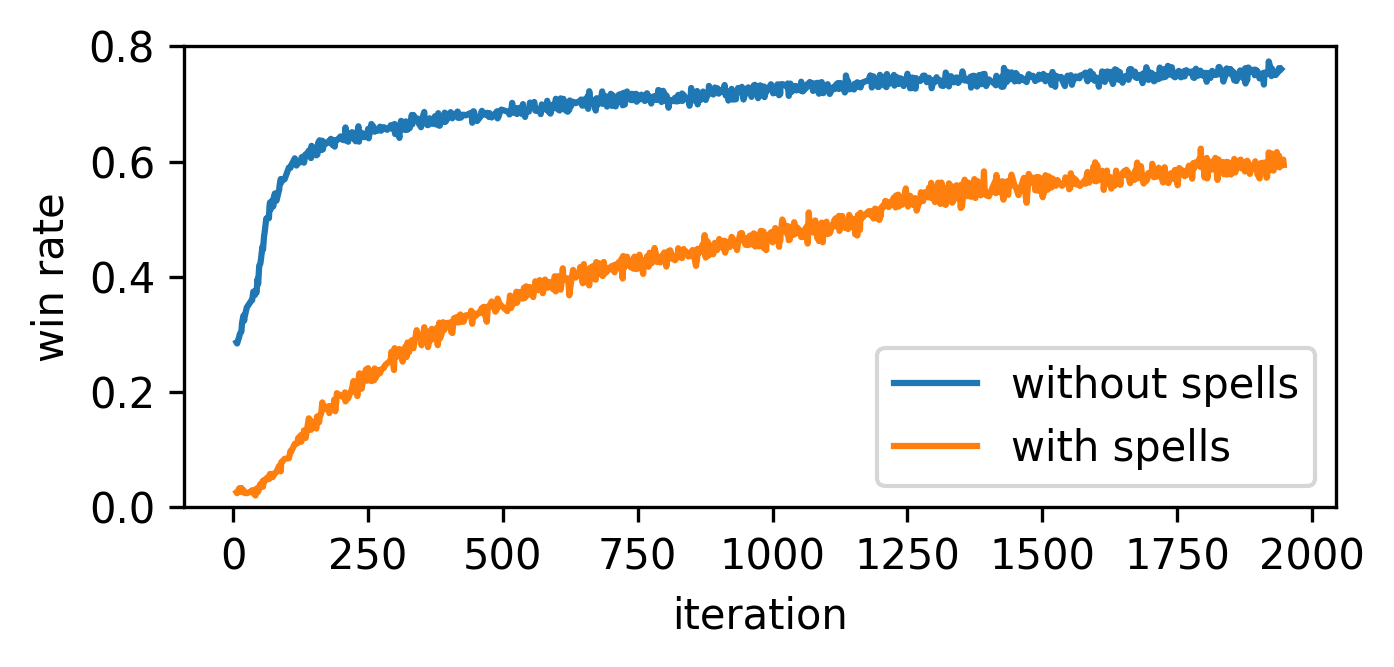}
    \includegraphics[width=0.8\columnwidth]
    {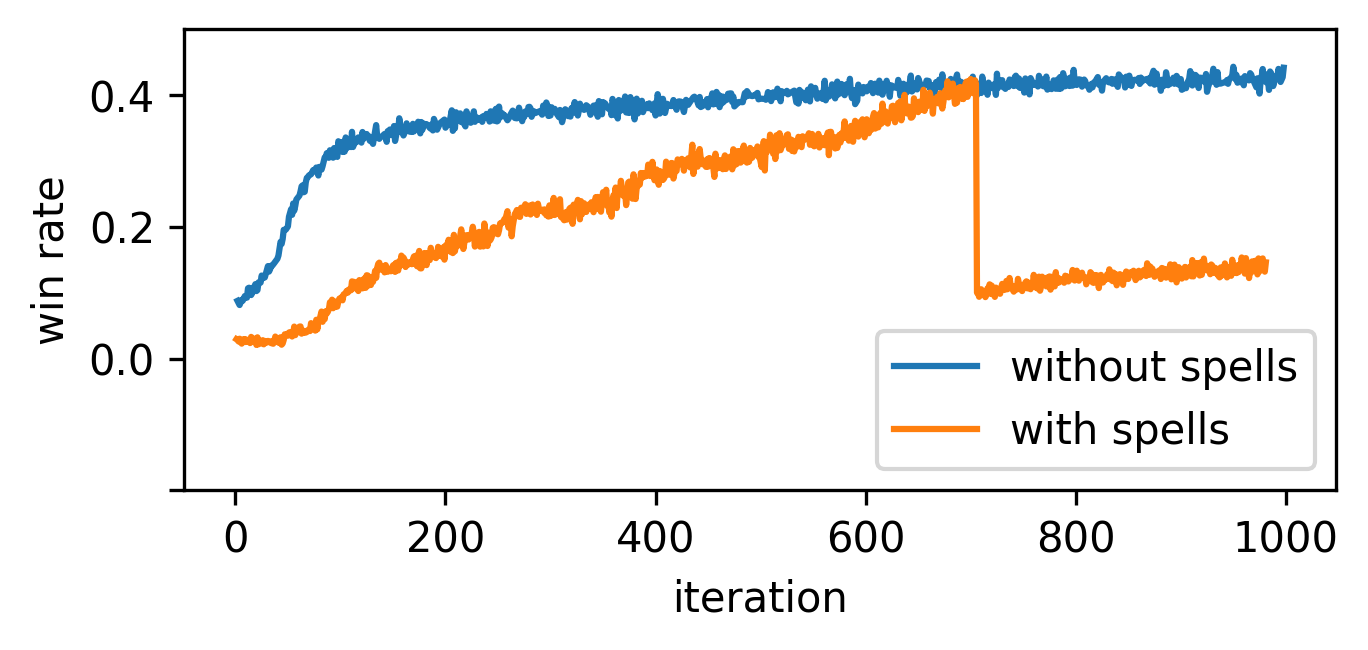}
\caption{Training without spells vs with spells.}\label{spells_no_spells}
\end{figure}

We think that this the result of the agents not learning to play no-op actions optimally, 
while no-op actions do not appear irrelevant as in the no-spell experiments. 
We hypothesize that this is one of the potential reasons for the gap in performance between the no-spell and spell agents, which we present in Figure~\ref{spells_no_spells}. In the upper
part of the figure we demonstrate the win rate of two agents
trained against the rule-based AI. In the same number of epochs the no-spell
agent obtains around $75\%$ win rate, while the spell agent reaches around $60\%$. 
The difference is even more apparent when
training against the stronger tree-search AI, see the lower part of
Figure~\ref{spells_no_spells}. There we present the win rate curve of an agent playing
without spells and an agent playing with spells, additionally pretrained against the
rule-based AI. We applied pretraining for the latter agent, since
otherwise the reward signal is not strong enough to initiate any learning. 
Note that the drop in win rate, which happens at the end pretraining, is a result
of stronger performance of the tree-search AI.

\section{Future Work}

Although the presented agents offer a good level of competition, the strategies
they learn lack long-term planning and the earlier states are not utilized,
since they use an MLP policy.  We experimented with providing past information
by stacking previous observations into a tensor with multiple channels,
similarly as in \cite{silver2017mastering}. However, we did not observe any
immediate benefit. Still, we think that endowing the agents with memory
could be a major improvement. Thus, in future work we would like to run experiments
with agents equipped with LSTM based policies.

We also plan to scale the agents further. We would like to augment the set of
available decks in order to make the policy more robust. We think that the
training procedure should also be scaled, by experimenting with more competing
agents. It would be interesting to adopt a bigger league of policies together
with exploiters or equilibrium strategies \cite{vinyals2019grandmasterli}.

Finally, we wish to tackle the performance gap between
spell and no-spell agents. In future work we would like to employ learning the no-op duration \cite{vinyals2019grandmasterli} together with recurrent policies and extended training, in order to mitigate this issue.

\section{Acknowledgments}

The authors would like to thank Marko Antonic, for providing detailed game
description, Sandra Tanackovic, for proofreading the paper and Marko Knezevic,
for valuable feedback during writing of this paper.

\bibliographystyle{aaai}
\bibliography{refs}

\end{document}